\documentclass[12pt]{article}
\usepackage[margin=0.9in]{geometry}

\usepackage[utf8]{inputenc} 
\usepackage[T1]{fontenc}    
\usepackage{hyperref}       
\usepackage{url}            
\usepackage{booktabs}       
\usepackage{amsfonts}       
\usepackage{nicefrac}       
\usepackage{microtype}      
\usepackage{xcolor}         

\usepackage{amsmath,amssymb,amsthm,eucal}
\usepackage[linesnumbered,ruled]{algorithm2e}
\usepackage{xspace}

\newcommand{\R}{\mathbb{R}}
\newcommand{\loss}{\mathcal{L}}

\DeclareMathOperator*{\argmax}{argmax}
\DeclareMathOperator*{\argmin}{argmin}

\usepackage{array}   
\newcolumntype{L}{>{$}l<{$}} 

\usepackage{graphicx}
\usepackage{subcaption}

\usepackage{hyperref}
\usepackage[capitalize,noabbrev]{cleveref}

\newcommand{\papertitle}{Boosting gets full Attention for Relational Learning}

\usepackage{arydshln}

\title{\papertitle}
\author{
  Mathieu Guillame-Bert \qquad Richard Nock \\
 Google\\
{\normalsize $\{$gbm,richardnock$\}$@google.com} \\
}
\begin{document}

\date{}

\maketitle

\begin{abstract}
More often than not in benchmark supervised ML, tabular data is flat, \textit{i.e.} consists of a single $m \times d$ (rows, columns) file, but cases abound in the real world where observations are described by a set of tables with structural relationships. Neural nets-based deep models are a classical fit to incorporate general topological dependence among description features (pixels, words, etc.), but their suboptimality to tree-based models on tabular data is still well documented. In this paper, we introduce an attention mechanism for structured data that blends well with tree-based models in the training context of (gradient) boosting. Each aggregated model is a tree whose training involves two steps: first, simple tabular models are learned descending tables in a top-down fashion with boosting's class residuals on tables' features. Second, what has been learned progresses back bottom-up via attention and aggregation mechanisms, progressively crafting new features that complete at the end the set of observation features over which a single tree is learned, boosting's iteration clock is incremented and new class residuals are computed. Experiments on simulated and real-world domains display the competitiveness of our method against a state of the art containing both tree-based and neural nets-based models.

\end{abstract}

\section{Introduction}

It is hard to exagerate the importance that structured data still has today in the industry \cite{big_data_2014, big_data_2_202}. Quite remarkably, while the unstructured data that feed deep learning methods gets undoubtedly more and more complex and rich in nature by the day \cite{DBLP:journals/corr/abs-2301-04856}, when it comes to structured data, machine learning (ML) research on (non-relational) tabular data~\cite{breiman_2001,friedman_2001} is significantly more popular than that on relational data. We believe that this disconnect can be partly attributed to the difficulty in obtaining publicly available, large, and representative benchmark relational datasets, as well as the effectiveness of hand-crafting tabular features from relational data (a.k.a. propositionalization) followed by the use of off-the-shelf tabular ML algorithms~\cite{soccer_2019}. 

There is however a singular phenomenon observed on tabular data: despite all their sophistication,  numerous attempts and the unrivalled sophistication of its models, deep learning is still mostly outperformed by simple, tree-based models on supervised learning tabular data \cite{govWD,DBLP:journals/corr/abs-2110-01889}. A key difference between deep models and tree-based models is that the former progressively leverage a natural topological dependence among description features (such as pixels in an image, words in a text, etc.), while the latter rely on probabilistic dependences between features, completely disregarding the topological component.

When it incorporates relations, structured data \textit{does} include a meaningful topological component: the relationships between the tables defining an observation, as exemplified in Figure \ref{fig:schema} below. Given the state of the art in deep learning and tree-based models, one would imagine that the design of accurate models for structured data would leverage what makes deep learning work on basis models defined by tree-based components. To our knowledge, no such models exists. One reason for this could be an observable strong divide between communities: from the deep learning side, tabular data looks like a "data bastion" still to be taken over with its models and architectures \cite{hdkoyTR,rDL} (and for a compact summary, \cite[Section 2, par. 2]{lcsbbgggTL}). There are also technical reasons for the hardness to bring tree-based models to the realm of deep learning, chief among which their non-differentiability \cite{hmehTA}. Finally, another technical hardness comes from the potential richness of schemata used to describe relational data, which can include tricky components like loops, self-loops, multiple semantically distinct relations between the same tables, etc. .

\textbf{Our paper} is a proposal, not just for the design of tree-based models exploiting the topological component of structured data, but also on how to train those in the algorithmic context that has been early coined as leading to the best off-the-shelf algorithms for tabular data: boosting \cite{fhtAL}. 

Training a single model operates in two distinct steps. In a first stage, simple regressive models are learned in a top-down fashion with respect to the tables' structure, and propagate a boosting's residuals used as label for the regressive models underneath. In a second stage, a bottom-up process progressively leverages the models learned using an attention mechanism that bears conceptual similarities to the one developed in transformers \cite{DBLP:conf/nips/VaswaniSPUJGKP17} as it leverages the importance of tables' features using a topology defined from relationships between tables\footnote{In transformers, topology arises naturally from the sequence of words in a text.}. At the end of the bottom-up process, the complete set of features, including those leveraged by attention and created by aggregation, is used to learn a tree model, after which boosting's next iteration starts (computing a leveraging coefficient for the current model, updating class residuals).

Our method leverages the gradient boosted learning framework~\cite{friedman_2001} and has a number of distinct features: our attention mechanism does not require differentiability, the whole algorithm is quick to run, has few hyper-parameters, with default values that perform well (no new hyper-parameters are introduced), does not necessitate feature pre-processing or normalization, and exhibits strong sample efficiency. Moreover, the method leverages a vast corpus of existing works, as it supports classification, regression, and ranking with ease. Despite the non-direct interpretability of gradient boosted trees, interpretative techniques like plotting trees, feature importance, and feature correlation can be applied to our method. Additionally, methods employed to interpret deep-learning attention models can also be extended to our method.

The next Section summarizes related work. Section \ref{sec:dommod} presents relational data, the attention mechanism and the way all the information gets integrated in the models learned. Section \ref{sec-learning} details our training algorithm and Section \ref{sec:experimental_results} presents experiments. Two last section discuss and conclude this paper with avenues for future work.

\section{Related work}

When it comes to relational data and observations described by multiple tables, one can have the option to just flatten the data and thus eventually end up with a single table akin to a conventional UCI domain\footnote{\url{https://archive.ics.uci.edu/ml/index.php}} but at the expense of a potential explosion in size, loss of structure, negative impact of duplicated features when noisy, etc. . At the age of deep learning, it is also fair to argue that the momentum should be on models being able to natively process structured data rather than data simplification to fit to a specific model's input. Of course, such models do exist. In the case of structured data, traditional deep neural networks could be a candidate: they progressively leverage the topological dependence among description features (pixels, words, nodes), with \textit{e.g.} Graph Neural Networks (GNN) leveraging general graph-representable relations \cite{gnn_2020}. Inductive Logic Programming (ILP) relies on logic based knowledge representation \cite{topdown_1998,fio_2005,ilp_1991,structural_tree_1991}, Relational Dependency Networks leverage logic relations in graph-based models \cite{rdn_2012}.

Some methods are even able to automatically generate tabular features from relational data~\cite{arda_2020,join_2016}. The extraction and propagation of discriminant signals along the relational structure of data is the central issue~\cite{struyf2006efficient} in Relational Learning~\cite{raedt2008logical}. ILP~\cite{ilp_1991} methods adopt a top-down approach~\cite{topdown_1998} to tackle relational learning. However, the set of all possible clauses is too large to be enumerated, and this problem is referred to as the ``short-sightedness'' in ILP literature. As a result, ``short-sighted'' heuristics have to be employed~\cite{struyf2006efficient}, which may potentially miss some discriminant signals. In contrast, connectionist methods, such as Graphical Neural Networks~\cite{gnn_2020}, "diffuse the discriminative signal" using backpropagation along the graph structure (``hops'') or an attention mechanism~\cite{attention_gnn_2017}.

\section{Relational Data, Attention and Models}\label{sec:dommod}

\subsection{Domain description}
\label{sec:pb_statement}

A \emph{dataset schema} is a tuple $(T, \{ P_t \}_{t \in T}, \{ R_t \}_{t \in T})$, where $T$ is the set of tables, $\{P_t \}_{t \in T}$ is the set of propositional attributes of each table, and $\{ R_t \}_{t \in T}$ is the set of relational attributes of each table.
The target table of $r \in R_t$ is noted $\dot{r} \in T$. In a directed graph representation of a schema, relations are displayed with arcs and the target table of a relation is the table pointed by the head of its arc (Subsection \ref{sub-model}). It is not to be confounded with a label in ML. To make a parallel with the ``flat'' world that most ML is accustomed to in \textit{e.g.} most UCI domains, such datasets have zero relations and a simple table with all description features. The description of the schema can be put in simple terms like the header section of an \texttt{arff} file \cite{witten2005data}. Such datasets with the simplest schema are called \textit{tabular}.

A \emph{dataset instance} is an instantiation of the schema; it is represented as a tuple $( \{ \mathcal{I}_t \}, \{ \mathcal{P}_{p,x} \}, \{ \mathcal{R}_{r,x} \} )$, where $\mathcal{I}_t$ are the rows in table $t \in T$, $\mathcal{P}_{p,x}$ are the propositional attribute value of $p \in P_t$ for row $x \in \mathcal{I}_t$, and $\mathcal{R}_{r,x} \subset \mathcal{I}_{\dot{r}}$ are the relational attribute value of $r \in R_t$ for row $x \in \mathcal{I}_t$.
When non ambiguous, we note $\mathcal{P}_{p,x}$ as $p[x]$, and $\mathcal{R}_{r,x}$ as $r[x]$.
Without loss of generality, we consider a propositional attribute value to be numerical i.e. $p[x] \in \R$.
We restrict relational attributes to represent sets of instances.
Finally, there exists a so-called \emph{root table} $s \in T$ whose rows are labeled. The objective is to train a model to predict the label $y[x]$ of instances $x \in \mathcal{I}_s$. We aim to learn a model $F$, whose domain is the complete schema (it does not contain the label in the root table). To simplify notations, it predicts $F(x)$ on $x \in \mathcal{I}_s$ -- we do not need to use the complete schema formalism for the input notation in $F$: passing a row of $\mathcal{I}_s$ implicitly includes the rest of the instance. We learn $F$ my minimizing a total loss:
\begin{align}
\sum_{x \in \mathcal{I}_s} \loss( F(x), y[x]),
\label{eq:loss}
\end{align}
where $\loss$ is the loss function used (e.g., cross-entropy, mean-squared-error, etc.).

\begin{figure}[t]
\centering
\includegraphics[scale=1.2]{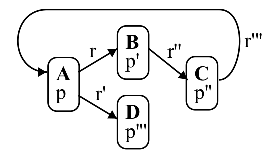}
\caption{Representation of the dataset schema with an oriented cycle.}\label{fig:schema}
\end{figure}

Figure~\ref{fig:schema} illustrates the notation introduced in this paper for a relational dataset consisting of four tables. Throughout the explanation of the algorithm, we will use this example as a reference.

In this example, the dataset contains four tables $T=\{A, B, C, D\}$, each with a single propositional attribute $P_A = \{p\}$, $P_B = \{p'\}$, $P_C = \{p''\}$, $P_D = \{p'''\}$. Table $A$ contains the label and has two relational attribute to $B$ and $D$ respectively i.e. $R_A = \{r, r'\}$ with $\dot{r} = B$ and $\dot{r}' = D$. Table $B$ as one relational attribute to $C$, and table $C$ has one relational attribute back to $A$. Table~\ref{tab:example} shows instances compatible with the schema.

\begin{table}[t]
\caption{Example of relational dataset whose schema is in Figure \ref{fig:schema}.}
\label{tab:example}
\center
\begin{tabular}{@{}LLLLLLLLL@{}}
\cmidrule(r){1-5} \cmidrule(l){7-9} 
A &  &  &  &  &  & B &  &  \\
id & p & r & r' & l &  & id & p' & r'' \\ \cmidrule(r){1-5} \cmidrule(l){7-9} 
a_0 & .1 & \{b_0\} & \{\} & 0 &  & b_0 & .3 & \{c_0, c_1\} \\
a_1 & .2 & \{b_0, b_1\} & \{d_0\} & 1 &  & b_1 & .4 & \{\} \\ \cmidrule(r){1-5} \cmidrule(l){7-9} 
 &  &  &  &  &  &  &  &  \\ \cmidrule(r){1-3} \cmidrule(lr){7-8}
C &  &  &  &  &  & D &  &  \\
id & p'' & r''' &  &  &  & id & p''' &  \\ \cmidrule(r){1-3} \cmidrule(lr){7-8}
c_0 & .5 & \{a_1\} &  &  &  & d_0 & .7 &  \\ \cmidrule(lr){7-8}
c_1 & .6 & \{\} &  &  &  &  &  &  \\ \cmidrule(r){1-3}
\end{tabular}
\end{table}

\begin{figure}[t]
\centering
\includegraphics[scale=1.2]{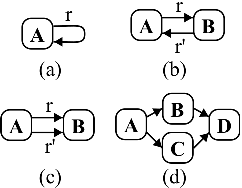}
\caption{Various basic patterns that can be represented with the dataset schema described in Section~\ref{sec:pb_statement}, and that can be aggregated to create complex schemata.}\label{fig:datasets_fun_schema}
\end{figure}

This data representation offers a large amount of flexibility.
Figure~\ref{fig:datasets_fun_schema} shows a few interesting examples. 
A dataset schema can contain a loop \textit{i.e.} a table can refer to itself as shown in Figure \ref{fig:datasets_fun_schema} (a) with $\dot{r} = t$ with $r \in R_t$, or a circuit \textit{via} a table that refers to itself though another table as shown in (b) with $\dot{r} = t'$ and $\dot{r'} = t$ with $r \in R_t$ and $r' \in R_t'$. In Figure \ref{fig:datasets_fun_schema} (d), a dataset schema contains \textit{non-oriented} circuits, \textit{i.e.} if we assimilate relationships to arcs and replace them by non directed edges, a cycle in the resulting graph. In Figure \ref{fig:datasets_fun_schema} (d), we have $\dot{r} = \dot{r}'$ with $r \in R_t$, $r' \in R_t'$ and $t \neq t'$.
Finally and importantly, two tables can have more than one relation as shown in Figure \ref{fig:datasets_fun_schema} (c). This is useful to express different types of relations between tables. 
As we already emphasized above, tabular datasets are a special case of a dataset that contains a single table and no relational attributes.

The main limitation of this schema representation is the absence of ordering in between related items. For example, if a row in table $A$ has a relation to two rows in table $B$, there is no order between those two rows for $A$. Though, this is interesting for further work. It is possible to bypass this limitation by encoding the position of the row as a feature, but this solution does not allow for (non-oriented) circuits. 

\subsection{Attention and Models}\label{sub-model}

A \emph{schedule} defines the information flow between the tables in a model.
A schedule $S$ is a rooted acyclic directed graph $S = (N, E)$ over the schema's topology. $N$ is the set of the nodes and $E$ the set of arcs, such that each node $n \in N$ maps to one table noted $\bar{n} \in T$, and each arc $e \in E$ maps to one relation noted $\bar{e} \in R_t$ with $e = (n, n') \in E \implies \bar{e} \in R_{\bar{n}}$.

\begin{figure}[t]
\centering
\includegraphics[scale=1.2]{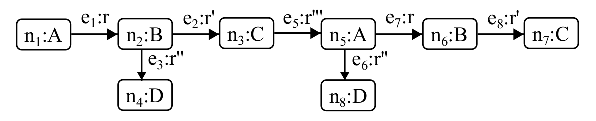}
\caption{A schedule compatible with the schema shown in Figure~\ref{fig:schema}.
Nodes are labelled as $n:t$ with $\bar{n} = t$.
Arcs are labelled as $e:r$ with $\bar{e} = r$.
$n_1$ is the root node i.e. $n_1 = u$ and $\bar{n}_1 = s = A$.
The schedule covers the schema twice: Each table is mapped by two nodes e.g. nodes $n_1$ and $n_5$ maps to table $A$.
}\label{fig:schedule}
\end{figure}

The \emph{root node} $u \in N$ maps to the root table $s$ i.e. $\bar{u} = s$. If the schema does not contain circuits, the schedule has the same topology as the schema \textit{i.e.} a one-to-one mapping between tables and nodes.
If the schema contains circuits, multiple nodes can map to the same table. In this case, we set the schedule to covers the schema a given number of time, set to 3 in all experiments.
Figure~\ref{fig:schedule} shows a schema compatible with the schema shown in Figure ~\ref{fig:schema}.

The proposed model is a gradient boosted model~\cite{friedman2000greedy} $F$ made of multiple ``weak'' models $\{h_{i, u} \}_i$:
\begin{align}
F(x) = \sum_i h_{i, u} ( B_{i,u}[x]).
\label{eq:model}
\end{align}
Weak models can be any supervised model. In our experiment, we use decision trees. As a standard gradient boosting model, the weak models are trained iteratively to minimize the empirical loss~(eq.\ref{eq:loss}).
The novelty of our method lies in the computation of the weak model input $B_{i,t}$.

The model input $B_{i,n}$ is is composed of four sub-terms:

\begin{eqnarray}
B_{i,n}[x] & = & B_{i,n}^{\textrm{prop}}[x] \oplus B_{i,n}^{\textrm{score}}[x] \label{eq:b_all} \oplus B_{i,n}^{\textrm{hard}}[x] \oplus B_{i,n}^{\textrm{soft}}[x], \label{eq:b_all}\\
B_{i,n}^{\textrm{prop}}[x] & = & \oplus_{p \in P_{\bar{n}}} p[x], \nonumber\\
  B_{i,n}^{\textrm{score}}[x] & = & \oplus_{e = (n, n') \in N} \phi( h_{i,n'}(B_{i,n'}[x']) ; \forall x' \in \bar{e}[x]), \nonumber\\
  B_{i,n}^{\textrm{soft}}[x] & = & \oplus_{e = (n, n') \in N; p \in P_{\bar{n'}}} \frac{\sum_{x' \in \bar{e}[x]} p[x'] h_{i,n'}(B_{i,n'}[x'])}{\sum_{x' \in \bar{e}[x]} h_{i,n'}(B_{i,n'}[x'])},  \nonumber\\
B_{i,n}^{\textrm{hard}}[x] & = & \oplus_{e = (n, n') \in N; p \in P_{\bar{n'}}} p\left[\argmax_{x' \in \bar{e}[x]} h_{i,n'}(B_{i,n'}[x']) \right], \nonumber
\end{eqnarray}
where $\oplus$ is the list  concatenation operator, and $\phi$ is an numerical aggregation operator. In our experiments, we use $\phi \in \{ \min, \max, \mathrm{mean}\}$. $B^{\textrm{prop}}$ are the propositional attribute values of $x$.
$B^{\textrm{score}}$ are the aggregated predictions of the child weak models in the schedule.
$B^{\textrm{soft}}$ are propositional attribute values of the child tables weighted by the child weak models predictions. This is a soft attention mechanism.
Finally, $B^{\textrm{hard}}$ are propositional attribute values of the child tables with the largest corresponding child weak models predictions. This is a hard attention mechanism.

To get an idea of what the schema formalism authorises ``above'' ML's classical tabular format, the whole model input would be reduced to $B^{\textrm{prop}}$, \textit{i.e.} the description variables, in that simple case.

\section{Learning}\label{sec-learning}

The model is trained using a series of gradient-boosting iterations. In each iteration $i$, a new weak model $h_{i,u}$ is trained on the root table $\dot{u} = s$ and subsequently incorporated into the ensemble model $F$.

To train each weak root model, the labels of the root table are utilized in conjunction with a specific loss function chosen for the prediction's objective (for example, a proper loss for binary classification, the mean-squared error loss for regression, etc.). Practically, every weak root model is trained to forecast the pseudo response that is calculated from the loss and the strong model as shown in eq.\ref{eq:loss}.

The pseudo response of models $\tilde{y}_{i,n'}$ on non-root tables is the pseudo response of their parent's model $\tilde{y}_{i,n}$ \emph{minus} the parent's model prediction $h_{i,n}[B^{\textrm{prop}}_{i,n}]$ as show in eq.~\ref{eq:non_root_label}. If a child table has multiple parents, its pseudo response is averaged over the parents. 
Note that Eq.~\ref{eq:b_all} establishes that the output of a model is computed using its \textit{children's} model predictions in the schedule. Thus, there is seemingly output information going both ways between tables during training. To disentangle the information flow, we propose to train each weak model twice.

During a \textit{forward pass}, nodes of the schedule are traversed according to the schedule topological ordering. For each node $n$, a weak model $h_{i,n}$ is trained without access to the predictions of weak models trained on children' outputs, that is $h_{i,n}$ is trained on $B_{i,n}^{\textrm{prop}}$ only.
The residual of the pseudo response of model $h_{i,n}$ defines the pseudo labels $\tilde{y}_{i,n'}$ of the weak models $h_{i,n'}$ with $(n,n') \in N$. Formally, we get:
\begin{eqnarray}
\tilde{y}_{i,n'}[x] & = &  \langle
\tilde{y}_{i,n}[x] - h_{i,n}[B^{\textrm{prop}}_{i,n}[x]] \rangle_{\mathcal{S}_n(x)},
\label{eq:non_root_label}\\
\mathcal{S}_n(x) & = & \{ (n,n') \in N : n' \in \bar{e}[x]\},\nonumber
\end{eqnarray}
with $\langle \cdot \rangle_{.}$ the set mean operator of the function in brackets using the set in index. Once all the weak models are trained, the $B_{i,n}^{\textrm{score}}$, $B_{i,n}^{\textrm{hard}}$, and $B_{i,n}^{\textrm{soft}}$ terms can be computed. Finally, all the weak models $h_{i,n}$ are retrained using $B_{i,n}$ instead of $B_{i,n}^{\textrm{prop}}$, and this happens during a \textit{backward pass}.

Algorithm~\ref{alg:full} details the training of the gradient-boosted model.

\begin{algorithm}
\SetAlgoLined
\DontPrintSemicolon
\caption{Training of the relational gradient boosted model.} \label{alg:full}
\KwData{$u$: a root node in a schedule,
$y[x]$: label for each instance in $x \in I_{\bar{u}}$,
$\loss$: The loss, $\gamma$: The shrinkage, $I$: The number of iterations.}
$F_0(x) = \argmin_\rho \sum_{x \in I_{ \dot{u}}} \loss(\rho, y[x])$ \tcp{initial predictions}
\For{$i \gets 1$ \KwTo $I$ } {
    $\tilde{y}_{i,u}[x] = \frac{\partial \loss(F_{i-1}[x], y[x])}{\partial F_{i-1}[x]}$ \tcp{pseudo label}
    \tcp{Forward pass}
    
    \For{$n \in N$ in topological order} {
        \If{$n \neq u$} {
            $\tilde{y}_{i,n}[x'] \gets$ residual pseudo labels from \eqref{eq:non_root_label} \;
        }
        $h_{i,n} \gets $ train weak model on $\tilde{y}_{i,n}$ and $B_{i,n}^{\textrm{prop}}$ \;
    }
    
    \tcp{Backward pass}
    
    \For{$n \in N$ in inverse topological order} {
        $h_{i,n} \gets $ train weak model on $\tilde{y}_{i,n}$ and $B_{i,n}$ \;
    }
    
    $F_i(x) = F_{i-1}(x) - \gamma h_{i,u}(x)$  \tcp{gradient boosting update}
}
\KwRet $F_I$ \;
\end{algorithm}

\section{Experimental Results}
\label{sec:experimental_results}

\begin{table}[t]
\centering
\caption{Statistics of the datasets used in the experiments (see text for explanations).}
\label{tab:datasets_statistics}
\resizebox{0.5\linewidth}{!}{\begin{tabular}{@{}lrrrr@{}}
\toprule
Dataset & \multicolumn{1}{l}{Rows} & \multicolumn{1}{l}{Connections} & \multicolumn{1}{l}{Features} & \multicolumn{1}{l}{Flattened} \\
  &   &   &  & \multicolumn{1}{l}{features} \\ \midrule
Synthetic & 44 802 & 53 745 & 4 & 10 \\
Financial & 1 063 259 & 18 439 & 15 & 146 \\
Mutagenesis & 15 567 & 25 862 & 8 & 57 \\
Arxiv & 169 343 & 2 332 484 & 128 & 384 \\
SST2 & 677 343 & 1 690 921 & 2 & 2 002 \\ \bottomrule
\end{tabular}}
\end{table}

\begin{figure}[t]
\centering
\includegraphics[scale=1.2]{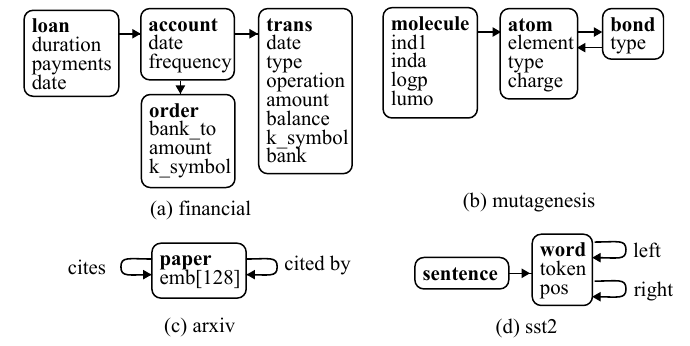}
\caption{Dataset schemata of the four real-world domains used in the experiments.}\label{fig:datasets_schema}
\end{figure}

The proposed algorithm is evaluated on a synthetic and several real-world relational datasets.
Table~\ref{tab:datasets_statistics} list datasets' key statistics. Figure~\ref{fig:datasets_schema} shows the schema of each dataset.

Decision forests algorithms shine by the quality of their out-of-the-box performance. That is, they generally perform well without hyperparameter tuning. For all our experiments, we have used the default parameters of the decision forests library as follows: maximum depth = 6, minimum number of observations in leaf = 5, number of iterations = 500, shrinkage = 0.1, and attributes sampling ratio = 0.2. These parameter values are inspired by the default parameters of the decision forests library used in our experiments~\cite{guillamebert2022yggdrasil}.

This section presents the datasets and respective evaluation protocols, describe the alternative evaluated methods, reports experimental performance of the proposed and alternative methods, and shows analyses providing insights into the model behavior.

\subsection{Datasets}

\subsubsection{Synthetic}

Our synthetic binary classification dataset is constructed in such a way that the discriminative information is embedded within the relational structure. This allows us to demonstrate how the proposed attention signal propagation approaches can effectively address the issue of relational shortsightedness.

The dataset's schema depicted in fig.~\ref{fig:schema}. The propositional features, namely $p$, $p'$, $p''$and $p'''$, are sampled uniformly and independently from the range $[0,1]$. The number of rows in relations $r$, $r'$, $r''$ are sampled uniformly from $\{0, 1, 2, 3\}$. Relation $r'''$ points to one item selected at random. The labels $y$ are computed from the following logical rule:

\begin{eqnarray}
    y[x] =
    \begin{cases}
      1 & \text{if } \exists
        y \in I_B, \exists z \in I_C : \left(\begin{array}{l}
        y \in r[x] \\
        \land z \in r''[y] \\
        \land p[x] \leq p''[z]
        \end{array}\right)\\
      0 & \text{otherwise}
    \end{cases}    
    \label{eq:synthetic_label}
\end{eqnarray}
Hence, an $A$ instance is labeled as $1$ only if it has one or more grandchildren $A \rightarrow B \rightarrow C$ with a value $p''$ greater than the $A$'s $p$. 59.80\% of the labels are negative. The simplicity of this dataset is just apparent as this dataset presents four critical challenges:

\begin{description}
    \item[Weak correlation:] A high value of $p$ is positively correlated with label $0$. Therefore, a propositional algorithm utilizing $p$ can predict the label better than random chance. Training a non-relational model using only $p$ leads to a maximum accuracy of 68.56\% as estimated numerically. Using the correct relational pattern, an accuracy of 100\% is attainable.
    
    \item[Wrong path]: The propositional attribute $p$ in isolation is insufficient to predict the label. Any pattern build solely on $p$ (e.g., $p[x] > 0.5$) is not a constituent of the ground truth pattern. Consequently, a constructive additive first-order logic (FOL) approach that commences with the literal $\mathrm{Greater}(p[x],0.5)$ is incapable of learning the optimal clause.
    
    \item[Shortsightedness trap] To identify the discriminant pattern, the model needs to extract the discriminant $p$ and $p''$ values and take them into account simultaneously. The pattern, represented in FOL notation, consists of three clauses shown in \eqref{eq:synthetic_label}, which may pose a challenge for inductive logic programming (ILP) algorithms.
    
    \item[Relational XOR] The first two clause $y \in I_B, z \in I_C ; y \in r[x] \land z \in r''[y]$, do not provide any immediate discriminative advantages for the model, so a short-sighted algorithm will not consider them. This scenario is analogous to the XOR problem in tabular learning, where no single feature nor a ``too simple'' model (linear separator) can beat random classification.
    
    \item[Diagonal pattern] The discriminant variables $p$ and $p''$ exhibit a linear threshold to decide the label, which may be challenging for axis-aligned decision trees to learn.
\end{description}

\subsubsection{Financial}

The financial dataset is a financial dataset used in the ECML PKDD 1999 competition. The dataset contains financial information about a set of clients. The original data is contained in 8 relational timestamped tables (load, order, transaction, account, district, client, disp and card). The objective is to predict the success of 682 loan using only information available at the time the load was created. As commonly done, we ignore the card, disp and client tables.
The algorithms are evaluated using a 20-fold cross-validation protocol.

\subsubsection{Mutagenesis}

The Mutagenesis dataset~\cite{debnath1991structure} is a collection of data about the mutagenicity of 230 chemical compounds. It contains information about the chemical structure of the compounds, as well as their toxicity. The objective is to predict the toxicity of compounds given their chemical structure.
The algorithms are evaluated using a 20-fold cross-validation protocol.

\subsubsection{Arxiv}

The Arxiv dataset~\cite{hu2021open} is a popular dataset to benchmark Graph Neural Networks~\cite{gnn}. The datset contains is a collection of 1.3M papers from the ArXiv preprint repository. Each paper is annotated with a 128-dimensional numerical vector computed from text embedding the title and abstract, and a subject areas (from 40 possibilities). Relational information, encoded as graphical edges, are used to encode 5.4M citations between papers. The objective is to predict the subject areas of a paper using citer papers.

The algorithm is evaluated using a train and test protocol using the OGGN train-validation-test split. Different evaluation protocol as commonly used on this dataset. In our experiments, the validation dataset is only used for early stopping. Test example features are not used during training. Early stopping appear non beneficial for our technique as the model never stop improving.

\subsubsection{Stanford Sentiment Treebank v2}

The Stanford Sentiment Treebank V2 (SST2)~\cite{sst2} is a widely used sentiment analysis text classification dataset. It consists of 68k sentences extracted from movie reviews and annotated with a sentiment label (positive or negative). The dataset is transformed into a relational format as follows. A \emph{sentence} table contains one row for each sentence and no propositional features. A \emph{word} table contains one row for each word in each sentence, linked to its corresponding sentence row. Each word row has two propositional features: the word token (categorical) and its position in the sentence (numerical). Additionally, the word table includes two relational features, \emph{left} and \emph{right}, which specify the preceding and following words for each word, respectively. The relational schema is illustrated in Figure \ref{fig:datasets_schema}.

\begin{table*}[t]
\centering
\caption{Accuracy of the candidate methods. The proposed method is noted ``Relational GBDT''. ``root features'' indicates that only features of the table containing the label are used. ``flatten features'' indicates that features from all the tables are flatten as described in Section \ref{sec:flatten}. No values are reported for root features of the SST2 dataset as the root table does not have features. Standard deviation is reported for cross-validated experiment.}
\label{tab:accuracies}
\begin{tabular}{lrrrrr}
\hline
\textbf{Method} & \textbf{synthetic} & \textbf{mutagenesis} & \textbf{financial} & \textbf{arxiv} & \textbf{sst2} \\ \hline
Relational GBDT (this paper) & \textbf{98.96} & \textbf{91.05} $\pm$ 6.25 & 88.33 $\pm$ 8.18 & \textbf{66.70} & \textbf{80.16} \\\hdashline
GBDT root features & 68.81 & 84.21 $\pm$ 8.15 & 86.67 $\pm$ 9.45 & 42.16 & \multicolumn{1}{l}{} \\
GBDT flatten features & 92.00 & 89.74 $\pm$ 6.55 & \textbf{88.41} $\pm$ 8.24 & 52.92 & 75.11 \\
RF root features & 68.49 & 84.21 $\pm$ 8.15 & 84.96 $\pm$ 7.44 & 44.56 & \multicolumn{1}{l}{} \\
RF flattent features & 91.68 & 89.74 $\pm$ 6.55 & 88.33 $\pm$ 9.16 & 55.91 & 52.06 \\
Linear root features & 67.80 & 89.47 $\pm$ 6.45 & 86.25 $\pm$ 8.55 & 52.48 & \multicolumn{1}{l}{} \\
Linear flatten features & 86.00 & 85.53 $\pm$ 7.42 & 81.97 $\pm$ 8.17 & 63.00 & 78.10 \\
Neural Net root features & 68.20 & 90.00 $\pm$ 6.96 & 85.42 $\pm$ 9.00 & 53.44 & \multicolumn{1}{l}{} \\
Neural Net flatten features & 87.20 & 86.58 $\pm$ 5.88 & 86.63 $\pm$ 7.95 & 58.96 & 74.31 \\ \hline
\end{tabular}
\end{table*}

\subsection{Alternative methods}
\label{sec:flatten}

A common approach for dealing with relational data is to convert it into tabular datasets through "flattening" or "proposionalization". For comparison purpose, we flatten all datasets and train and evaluate a collection of tabular models. The flattening is done using the same schedule as our method as follow. For each pair of nodes $(n, n') \in N$, numerical features in $n'$ are aggregated using the mean operator, and the values are propagated to $n$. For categorical features, the distribution of each unique value is propagated.
The number of generated (``flattened'') features is reported in Table~\ref{tab:datasets_statistics}.

The following tabular learning algorithms are considered: Random Forests and Gradient Boosted Trees models using Yggdrasil Decision Forests 0.1.0~\cite{ydf}. Linear models are trained using scikit-learn 1.4.0~\cite{scikit-learn}. Neural network models are computed using TensorFlow 2.15.0~\cite{abadi2016tensorflow}. When available, we use the default hyper-parameter value provided by each library. Neural network models are three layer models with 128 units each and with Relu activation functions.

For comparison purposes, each tabular learning algorithm is also applied to the table containing labels without flattening features, meaning that only the features in the table containing the label are used.

\subsection{Evaluation}

Table~\ref{tab:accuracies} shows the accuracy of our approach along side comparable works.
We observe that in all experiments except one (financial), the proposed method (Relational GBDT) outperforms all other evaluated methods. On the financial datasets, our method is tied for second place. What is perhaps more telling is the difference in accuracy to our closest contender: it exceeds $6\%$ on synthetic and $3\%$ on arxiv, while we are less than $.1\%$ away from the best performing method on financial. There is little doubt that our approach to design an attention mechanism for relational data indeed manages to leverage relational information if we compare it to GBDT and RF (almost $14\%$ improvement on \textbf{arxiv}).

\subsection{Model analysis}

\subsubsection{Interpretation of model on synthetic dataset}

\begin{figure*}
     \centering
     \resizebox{\textwidth}{!}{\begin{tabular}{cccc}
     \begin{subfigure}[b]{0.33\textwidth}
         \centering
         \includegraphics[width=\textwidth]{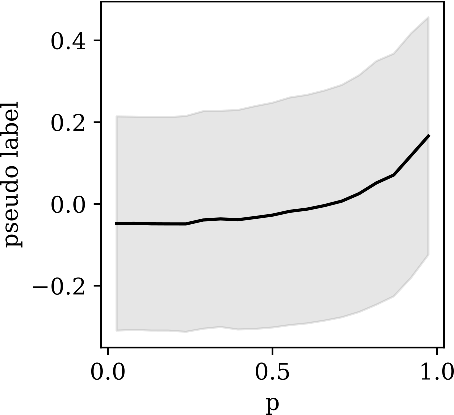}
         \caption{A's pseudo label on the forward pass}
         \label{fig:y equals x}
     \end{subfigure}
     &
     \begin{subfigure}[b]{0.33\textwidth}
         \centering
         \includegraphics[width=\textwidth]{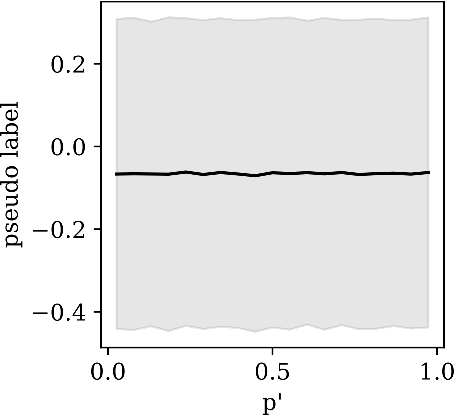}
         \caption{B's pseudo label on the forward pass}
         \label{fig:three sin x}
     \end{subfigure}
     &
     \begin{subfigure}[b]{0.33\textwidth}
         \centering
         \includegraphics[width=\textwidth]{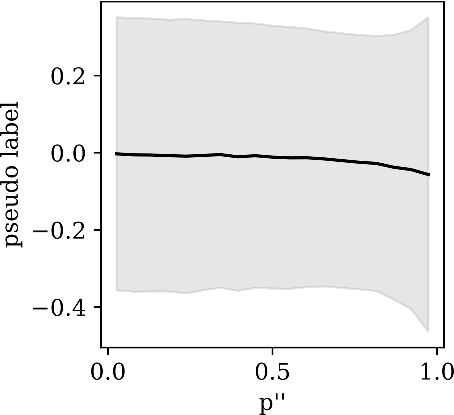}
         \caption{C's pseudo label on the forward pass}
         \label{fig:five over x}
     \end{subfigure}
     &
     \begin{subfigure}[b]{0.33\textwidth}
         \centering
         \includegraphics[width=\textwidth]{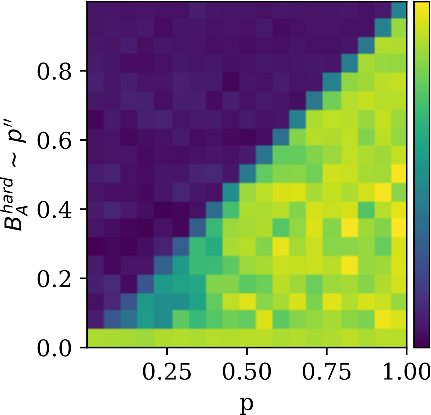}
         \caption{A's pseudo label on the backward pass}
         \label{fig:five over x}
     \end{subfigure}
     \end{tabular}}
        \caption{On the synthetic dataset, average and one standard deviation range of the pseudo-labels on tables $A$, $B$, and $C$, on the forward and backward pass of the first training iteration. During the forward pass (a, b and c), $B^{\textrm{prop}}$ contains a single feature for each table. On table $A$, feature $p$ is partially discriminative showing an upward trend relation with a large error margin. On table $B$, feature $p'$ is not discriminative showing no relation between with the pseudo label, which complies with the fact that it does not appear in the label definition \eqref{eq:synthetic_label}. On table $C$, feature $p''$ is partially discriminative with a slight downward trend relation. This shows that $p''$ is discriminative with respect to the residual of the pseudo label on table $B$, which is equal to the residual of the pseudo label on table $A$. Therefore, $p''$ can be selected by the hard attention mechanism $B^{\textrm{hard}}$ and forwarded to table $B$ and then $A$. During the backward pass on table $A$, feature $p$ and the hard attention $B^{\textrm{hard}}_A$ which is equal to the selected $p''$, are together highly discriminative showing very good prediction of $A$'s pseudo label (d) and a frontier that closely matches the domain's optimal separation, see \eqref{eq:synthetic_label}.}
        \label{fig:toy_step_by_step}
\end{figure*}

In this section, we look at the internal representation of the learning algorithm on the synthetic dataset. Figure~\ref{fig:toy_step_by_step} shows the pseudo-labels used to train the weak tabular models on tables A, B, and C during the first training iteration.

\setlength\tabcolsep{1pt}

Variable importances are a popular technique to analyze models and understand underlying systems.
By aggregating the variable importance of weak models for each table, we measure the importance of both propositional features and relational ones.
We use the average of all the paths of the minimal depth the feature is first used. This value is normalized between 0 and 1, where 1 being high importance and 0 being that the feature is never employed.
On the table $A$, the two most important features are the hard attention on the feature $p''$ (1.0 importance) and the propositional feature $p$ (0.62 importance), other attention features having much lower importance scores (0.38 importance for the top one).

\subsubsection{Interpretation of model on financial dataset}

Similarly as for the synthetic dataset, we can study the variable importance of the relational model on the financial dataset. On the \emph{loan} table containing the label, the most important feature is the \emph{day of the month} in \emph{date} from the \emph{transaction} table and extracted by two hops of hard attention through the \emph{account} table.
This indicates that information from transactions, including the bank used to make transactions, are critical. That a variable like \emph{day of the month} be important is not so surprising: it may factor additional domain contextual information about the client and the bank's processes that contribute to be informative to predict the label. 
To understand the how the hard attention on the transaction table work, we can look at the variable importance on the "transaction" table: We see that the most important feature is the transaction "bank" (0.67) followed by transaction "year" (0.53) and the transaction "balance" (0.29).
This illustrates how we can interpret relational model by unfolding the variable importances over the tables.
\section{Discussion}

\paragraph{Attention and (non-)differentiability} Our approach extends the gradient boosting framework~\cite{friedman_2001} to relational data by incorporating an attention mechanism. Specifically, we propose an learning algorithm that can extend any weak learner, such as a decision tree, linear model, or nearest neighbor, to support relational data.
Like Attention GNN~\cite{attention_gnn_2017}, the proposed method utilizes a signal diffusion process that incorporates an attention mechanism~\cite{transformer_2022}. However, unlike Transformer's attention, our method introduces a novel attention mechanism that does not necessitate model differentiability and can be employed with any weak learner such as one learning decision trees.

\paragraph{Attention and (non-)explainability} Attention is a mechanism that learns to put emphases within parts of models. It is hardcoded in our definition, and similarly for neural nets \cite{DBLP:conf/nips/VaswaniSPUJGKP17,attention_gnn_2017}. With the ever expanding size of models, such a mechanism offers the natural opportunity to interpret those important parts of the classification process, but it faces a major downside when the structure and classification of the classifier is not transparent enough, which is the case for neural nets \cite{DBLP:journals/corr/abs-2307-09050}. Another downside is the fact that interpretability is substantially subjective and so any automatic post-processing of a model must not induce an eventual user-dependent bias. This is where attention for relational data can kill two birds in one shot: first, unlike \textit{e.g.} text, a dataset schema is structured and built from domain-dependent human design, which can therefore integrate interpretability goals \textit{ex ante}, directly from its design. Second, each attribute in a schema is usually informative \textit{per se}, unlike \textit{e.g.} images where a pixel taken out of its context has zero informative content for the task at hand. In our context, attention singles out content from the schema that can be informative even disregarding the rest of it (see our Experiments).

\paragraph{Attention and boosting} The original boosting model shows how one can combine the outputs of moderately accurate classifiers into an ensemble which can be arbitrarily accurate \cite{kTO,kvAI}. There exists two main ways to get this ``boost'' in accuracy: by splitting data (as in decision trees) or by making linear combinations of classifiers \cite{mnwRC}. One then assumes that the split or the function added in the linear combination satisfies a so-called weak learning assumption, and these local ``edits'' to the final model get the boosting improvement as more and more are added. Attention, as we devise it, hints on the best part(s) of a model. In the worst case, it should not bring improvements in terms of boosting rates -- if each of these parts satisfies the weak learning assumption, so does the best one, but nothing more --. However, in a context where some of these parts may achieve more than the weak learning assumption, \textit{e.g.} as could be the case when exploiting the \textit{depth} of a schema and the classifiers involved in the backward pass of Algorithm \ref{alg:full}, some improvements could be observed. This entails the question of the optimal location for updating pseudo-labels (Step 3 in Algorithm \ref{alg:full}) and certainly deserves further... attention.
\section{Conclusion}

In this paper, we introduce an attention mechanism tailor made for relational data, which leverages dependencies between tables to locally elicit the part of attributes that are useful for classification. We also introduce a training algorithm in the gradient boosting framework which leverages attention through a two-pass mechanism to propagate boosting residuals and train classifiers. Experiments demonstrate the competitiveness of this algorithm against a variety of state of the art contenders and the opportunities that the attention mechanism brings for explainable AI in the context of relational data.
\section{Code availability}

Please contact Mathieu at gbm@google.com.

\bibliographystyle{plain}
\bibliography{references}

\begin{thebibliography}{10}

\bibitem{abadi2016tensorflow}
Martín Abadi, Paul Barham, Jianmin Chen, Zhifeng Chen, Andy Davis, Jeffrey
  Dean, Matthieu Devin, Sanjay Ghemawat, Geoffrey Irving, Michael Isard,
  Manjunath Kudlur, Josh Levenberg, Rajat Monga, Sherry Moore, Derek~G. Murray,
  Benoit Steiner, Paul Tucker, Vijay Vasudevan, Pete Warden, Martin Wicke, Yuan
  Yu, and Xiaoqiang Zheng.
\newblock Tensorflow: A system for large-scale machine learning, 2016.

\bibitem{DBLP:journals/corr/abs-2301-04856}
Cem Akkus, Luyang Chu, Vladana Djakovic, Steffen Jauch{-}Walser, Philipp Koch,
  Giacomo Loss, Christopher Marquardt, Marco Moldovan, Nadja Sauter, Maximilian
  Schneider, Rickmer Schulte, Karol Urbanczyk, Jann Goschenhofer, Christian
  Heumann, Rasmus Hvingelby, Daniel Schalk, and Matthias A{\ss}enmacher.
\newblock Multimodal deep learning.
\newblock {\em CoRR}, abs/2301.04856, 2023.

\bibitem{big_data_2014}
Rudolph~T. Bedeley.
\newblock Big data opportunities and challenges: The case of banking industry,
  2014.

\bibitem{topdown_1998}
Hendrik Blockeel and Luc {De Raedt}.
\newblock Top-down induction of first-order logical decision trees.
\newblock {\em Artificial Intelligence}, 101(1):285--297, 1998.

\bibitem{DBLP:journals/corr/abs-2110-01889}
Vadim Borisov, Tobias Leemann, Kathrin Se{\ss}ler, Johannes Haug, Martin
  Pawelczyk, and Gjergji Kasneci.
\newblock Deep neural networks and tabular data: {A} survey.
\newblock {\em IEEE T. Neural Networks and Learning Systems}, 2023.

\bibitem{breiman_2001}
L~Breiman.
\newblock Random forests.
\newblock {\em Machine Learning}, 45:5--32, 10 2001.

\bibitem{arda_2020}
Nadiia Chepurko, Ryan Marcus, Emanuel Zgraggen, Raul~Castro Fernandez, Tim
  Kraska, and David~R. Karger.
\newblock {ARDA:} automatic relational data augmentation for machine learning.
\newblock {\em Proc. {VLDB} Endow.}, 13(9):1373--1387, 2020.

\bibitem{raedt2008logical}
Luc De~Raedt.
\newblock {\em Logical and relational learning}.
\newblock Springer Science \& Business Media, 2008.

\bibitem{debnath1991structure}
Asim~Kumar Debnath, Rosa~L Lopez~de Compadre, Gargi Debnath, Alan~J Shusterman,
  and Corwin Hansch.
\newblock Structure-activity relationship of mutagenic aromatic and
  heteroaromatic nitro compounds. correlation with molecular orbital energies
  and hydrophobicity.
\newblock {\em Journal of medicinal chemistry}, 34(2):786--797, 1991.

\bibitem{big_data_2_202}
Diane Dolezel and Alexander~J. McLeod.
\newblock Big-data skills: Bridging the data science theory-practice gap in
  healthcare.
\newblock {\em Perspectives in health information management}, 18 Winter, 2021.

\bibitem{fhtAL}
J.~Friedman, T.~Hastie, and R.~Tibshirani.
\newblock {A}dditive {L}ogistic {R}egression : a {S}tatistical {V}iew of
  {B}oosting.
\newblock {\em Ann. of Statistics}, 28:337--374, 2000.

\bibitem{friedman2000greedy}
Jerome~H. Friedman.
\newblock Greedy function approximation: A gradient boosting machine.
\newblock {\em Annals of Statistics}, 29:1189--1232, 2000.

\bibitem{friedman_2001}
Jerome~H. Friedman.
\newblock {Greedy function approximation: A gradient boosting machine.}
\newblock {\em The Annals of Statistics}, 29(5):1189 -- 1232, 2001.

\bibitem{govWD}
L.~Grinsztajn, E.~Oyallon, and G.~Varoquaux.
\newblock Why do tree-based models still outperform deep learning on tabular
  data?
\newblock In {\em NeurIPS'22 Datasets and Benchmarks}, 2022.

\bibitem{guillamebert2022yggdrasil}
Mathieu Guillame-Bert, Sebastian Bruch, Richard Stotz, and Jan Pfeifer.
\newblock Yggdrasil decision forests: A fast and extensible decision forests
  library, 2022.

\bibitem{ydf}
Mathieu Guillame-Bert, Sebastian Bruch, Richard Stotz, and Jan Pfeifer.
\newblock Yggdrasil decision forests: A fast and extensible decision forests
  library.
\newblock In {\em Proceedings of the 29th ACM SIGKDD Conference on Knowledge
  Discovery and Data Mining}, KDD '23, page 4068–4077, New York, NY, USA,
  2023. Association for Computing Machinery.

\bibitem{gnn}
William~L. Hamilton, Rex Ying, and Jure Leskovec.
\newblock Representation learning on graphs: Methods and applications.
\newblock {\em CoRR}, abs/1709.05584, 2017.

\bibitem{hmehTA}
Noah Hollmann, Samuel M{\"u}ller, Katharina Eggensperger, and Frank Hutter.
\newblock {TabPFN}: a transformer that solves small tabular classification
  problems in a second.
\newblock In {\em NeurIPS'22 Table Representation Learning Workshop}, 2022.

\bibitem{hu2021open}
Weihua Hu, Matthias Fey, Marinka Zitnik, Yuxiao Dong, Hongyu Ren, Bowen Liu,
  Michele Catasta, and Jure Leskovec.
\newblock Open graph benchmark: Datasets for machine learning on graphs, 2021.

\bibitem{soccer_2019}
Ond\v{r}ej Hub\'{a}\v{c}ek, Gustav \v{S}ourek, and Filip \v{Z}elezn\'{y}.
\newblock Learning to predict soccer results from relational data with gradient
  boosted trees.
\newblock {\em Machine Learning}, 108(1):29–47, jan 2019.

\bibitem{hdkoyTR}
Madelon Huselbos, Haoyu Dong, Bojan Karlas, Laurel Orr, and Pengcheng Yin.
\newblock {Table Representation Learning Workshop at NeurIPS}, 2022.
\newblock
  \url{https://table-representation-learning.github.io/organization.html}.

\bibitem{kvAI}
M.~J. Kearns and U.~V. Vazirani.
\newblock {\em An {I}ntroduction to {C}omputational {L}earning {T}heory}.
\newblock M.I.T. Press, 1994.

\bibitem{kTO}
M.J. Kearns.
\newblock Thoughts on hypothesis boosting, 1988.
\newblock ML class project.

\bibitem{join_2016}
Arun Kumar, Jeffrey Naughton, Jignesh~M. Patel, and Xiaojin Zhu.
\newblock To join or not to join? thinking twice about joins before feature
  selection.
\newblock In {\em Proceedings of the 2016 International Conference on
  Management of Data}, SIGMOD '16, page 19–34, New York, NY, USA, 2016.
  Association for Computing Machinery.

\bibitem{lcsbbgggTL}
Roman Levin, Valeriia Cherepanova, Avi Schwartzschild, Arpit Bansal, {C. Bayan
  Bruss}, Tom Goldstein, Andrew~Gordon Wilson, and Micah Goldblum.
\newblock Transfer learning with deep tabular models.
\newblock In {\em NeurIPS'22 Table Representation Learning Workshop}, 2022.

\bibitem{transformer_2022}
Tianyang Lin, Yuxin Wang, Xiangyang Liu, and Xipeng Qiu.
\newblock A survey of transformers.
\newblock {\em AI Open}, 3:111--132, 2022.

\bibitem{mnwRC}
Y.~Mansour, R.~Nock, and R.-C. Williamson.
\newblock Random classification noise does not defeat all convex potential
  boosters irrespective of model choice.
\newblock In {\em ICML'23}, 2023.

\bibitem{fio_2005}
Drew Mellor.
\newblock A first order logic classifier system.
\newblock In {\em Proceedings of the 7th Annual Conference on Genetic and
  Evolutionary Computation}, GECCO '05, page 1819–1826, New York, NY, USA,
  2005. Association for Computing Machinery.

\bibitem{ilp_1991}
Stephen~H. Muggleton.
\newblock Inductive logic programming.
\newblock {\em New Generation Computing}, 8:295--318, 1991.

\bibitem{rdn_2012}
Sriraam Natarajan, Tushar Khot, Kristian Kersting, Bernd Gutmann, and Jude
  Shavlik.
\newblock Gradient-based boosting for statistical relational learning: The
  relational dependency network case.
\newblock {\em Machine Learning}, 86:25--56, 05 2012.

\bibitem{DBLP:journals/corr/abs-2307-09050}
Yingjie Niu, Ming Ding, Maoning Ge, Robin Karlsson, Yuxiao Zhang, and Kazuya
  Takeda.
\newblock R-cut: Enhancing explainability in vision transformers with
  relationship weighted out and cut.
\newblock {\em CoRR}, abs/2307.09050, 2023.

\bibitem{scikit-learn}
F.~Pedregosa, G.~Varoquaux, A.~Gramfort, V.~Michel, B.~Thirion, O.~Grisel,
  M.~Blondel, P.~Prettenhofer, R.~Weiss, V.~Dubourg, J.~Vanderplas, A.~Passos,
  D.~Cournapeau, M.~Brucher, M.~Perrot, and E.~Duchesnay.
\newblock Scikit-learn: Machine learning in {P}ython.
\newblock {\em Journal of Machine Learning Research}, 12:2825--2830, 2011.

\bibitem{rDL}
Sebastian Raschka.
\newblock A short chronology of deep learning for tabular data, 2022.
\newblock
  \url{https://sebastianraschka.com/blog/2022/deep-learning-for-tabular-data.html}.

\bibitem{sst2}
Richard Socher, Alex Perelygin, Jean Wu, Jason Chuang, Christopher~D. Manning,
  Andrew Ng, and Christopher Potts.
\newblock Recursive deep models for semantic compositionality over a sentiment
  treebank.
\newblock In David Yarowsky, Timothy Baldwin, Anna Korhonen, Karen Livescu, and
  Steven Bethard, editors, {\em Proceedings of the 2013 Conference on Empirical
  Methods in Natural Language Processing}, pages 1631--1642, Seattle,
  Washington, USA, October 2013. Association for Computational Linguistics.

\bibitem{struyf2006efficient}
Jan Struyf, Jesse Davis, and David Page.
\newblock An efficient approximation to lookahead in relational learners.
\newblock In {\em European conference on machine learning}, pages 775--782.
  Springer, 2006.

\bibitem{DBLP:conf/nips/VaswaniSPUJGKP17}
Ashish Vaswani, Noam Shazeer, Niki Parmar, Jakob Uszkoreit, Llion Jones,
  Aidan~N. Gomez, Lukasz Kaiser, and Illia Polosukhin.
\newblock Attention is all you need.
\newblock In {\em Advances in Neural Information Processing Systems 30: Annual
  Conference on Neural Information Processing Systems 2017, December 4-9, 2017,
  Long Beach, CA, {USA}}, pages 5998--6008, 2017.

\bibitem{attention_gnn_2017}
Petar Veličković, Guillem Cucurull, Arantxa Casanova, Adriana Romero, Pietro
  Liò, and Yoshua Bengio.
\newblock Graph attention networks, 2017.

\bibitem{structural_tree_1991}
Larry Watanabe and Larry Rendell.
\newblock Learning structural decision trees from examples.
\newblock In {\em Proceedings of the 12th International Joint Conference on
  Artificial Intelligence - Volume 2}, IJCAI'91, page 770–776, San Francisco,
  CA, USA, 1991. Morgan Kaufmann Publishers Inc.

\bibitem{witten2005data}
Ian~H Witten and Eibe Frank.
\newblock {\em Data Mining: Practical machine learning tools and techniques}.
\newblock Morgan Kaufmann, 2005.

\bibitem{gnn_2020}
Jie Zhou, Ganqu Cui, Shengding Hu, Zhengyan Zhang, Cheng Yang, Zhiyuan Liu,
  Lifeng Wang, Changcheng Li, and Maosong Sun.
\newblock Graph neural networks: A review of methods and applications.
\newblock {\em AI Open}, 1:57--81, 2020.

\end{thebibliography}

\end{document}